\documentclass{article}
\usepackage{spconf,amsmath,epsfig}

\let\OLDthebibliography\thebibliography
\renewcommand\thebibliography[1]{
  \OLDthebibliography{#1}
  \setlength{\parskip}{0pt}
  \setlength{\itemsep}{0pt plus 0.3ex}
}
\usepackage{paralist}
\usepackage{rotating}
\usepackage{url}

\begin{document}\sloppy

% Example definitions.
% --------------------
\def\eg{{\em e.g.}}
\def\ie{{\em i.e.}}

% Title.
% ------
\title{DeepFake Detection: Current Challenges and Next Steps}
%
% Single address.
% ---------------
\name{Siwei Lyu}
\address{Computer Science Department \\University at Albany, State University of New York}

\maketitle

\begin{abstract}
High quality fake videos and audios generated by AI-algorithms (the deep fakes) have started to challenge the status of videos and audios as definitive evidence of events. In this paper, we highlight a few of these challenges and discuss the research opportunities in this direction.
\end{abstract}
\section{Introduction}
\label{sec:intro}

Falsified videos created by AI algorithms, in particular, deep neural networks (DNNs), are a recent twist to the disconcerting problem of online disinformation. Although fabrication and manipulation of digital images and videos are not new \cite{farid08}, the rapid development of DNNs in recent years has made the process to create convincing fake videos increasingly easier and faster. DNN generated fake videos first caught the public's attention in late 2017, when a Reddit account with name {\em Deepfakes} began posting synthetic pornographic videos generated using a DNN-based face-swapping algorithm. Subsequently, the term DeepFake have been used more broadly to refer to any AI generated impersonating videos. 

Currently, there are three major types of DeepFake videos.
\begin{itemize}
    \item Head puppetry entails synthesizing a video of a target person’s whole head and upper-shoulder using a video of a source person’s head, so the synthesized target appears to behave the same way as the source.
    \item Face swapping involves generating a video of the target with the faces replaced by synthesized faces of the source while keeping the same facial expressions.
    \item Lip syncing is to create a falsified video by only manipulating the lip region so that the target appears to speak something that s/he does not speak in reality. 
\end{itemize}
\begin{figure}[t]
    \centering
    \includegraphics[width=.5\textwidth]{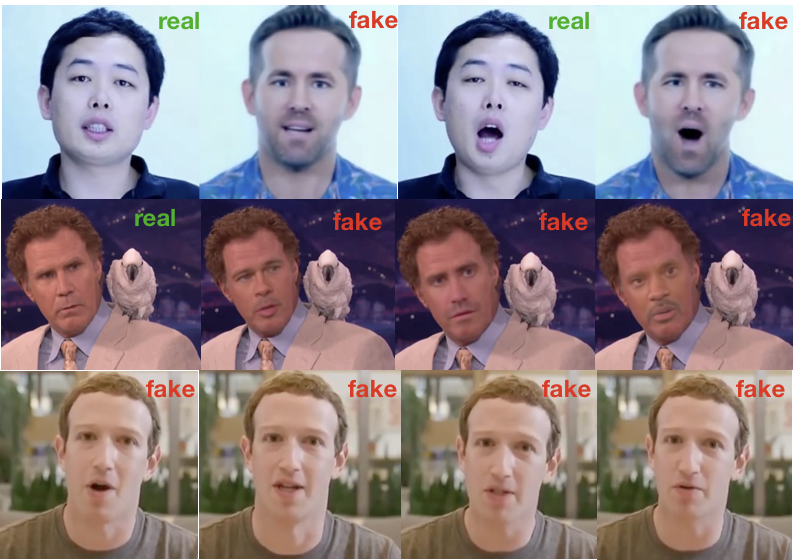}
    ~\vspace{-2em}
    \caption{\em \small Examples of DeepFake videos: (top) Head puppetry, (middle) face swapping, and (bottom) lip syncing. }
    \label{fig:example}
    ~\vspace{-2em}
\end{figure}
Figure \ref{fig:example} shows some example frames of each type of DeepFake videos aforementioned. As the first examples of DeepFakes, face swapping has been commercialized and mainstreamed through readily available software freely available on GitHub, \eg, {\tt FakeApp} \cite{fakeapp}, {\tt DFaker} \cite{DFaker}, {\tt faceswap-GAN} \cite{faceswap-gan}, {\tt faceswap} \cite{faceswap}, and {\tt DeepFaceLab} \cite{DeepFaceLab}. 
There are also emerging online services that can generate DeepFake videos on demand (\url{https://deepfakesweb.com}), and there are many online discussion fora on DeepFakes. Furthermore, several start-up companies also commercialized tools that can potentially be used to make DeepFakes, such as Synthesia\footnote{\url{https://www.synthesia.io/}.} and Canny AI\footnote{\url{https://www.cannyai.com/}.}.

While there are interesting and creative applications of the DeepFake videos, due to the strong association of faces to the identity of an individual, they can also be weaponized. Well-crafted DeepFake videos can create illusions of a person's presence and activities that do not occur in reality, which can lead to serious political, social, financial, and legal consequences \cite{survey_chesney_citron_2018}. The potential threats range from revenge pornographic videos of a victim whose face is synthesized and spliced in, to realistically looking videos of state leaders seeming to make inflammatory comments they never actually made, a high-level executive commenting about her company's performance to influence the global stock market, or an online sex predator masquerades visually as a family member or a friend in a video chat. The high stakes spawn wide media coverage of this topic in the past two years, and the US congress has had two public hearings to this problem.

With the escalated concerns over DeepFakes, there is a surge of interest in developing DeepFake detection methods with 
significant progress witnessed in the past two years. This includes (1) a slew of effective detection methods developed in less than two years, mostly based on deep learning \cite{afchar2018mesonet,guera2018deepfake,li2018ictu,yang2018exposing,matern2019exploiting,li2019exposing,sabir2019recurrent,roessler2019faceforensics++,nguyen2019capsule,nguyen2019multi,nguyen2019capsulev2}; (2) the availability of several large-scale DeepFake video datasets \cite{yang2018exposing,korshunov2018deepfakes,roessler2019faceforensics++,DDD_GoogleJigSaw2019,dolhansky2019deepfake}; and (3) two public challenges dedicated to DeepFake detection, namely, the DARPA MFC18 Synthetic Data Detection Challenge and the Facebook {\em DeepFake Detection Challenge}\footnote{\url{https://deepfakedetectionchallenge.ai}.}.

Notwithstanding this progress, there are a number of critical problems that are yet to be resolved for existing DeepFake detection methods. Furthermore, in the foreseeable future, it is expected that the generation of DeepFake videos will continue evolving, it is thus important to anticipate such new developments and improve the detection methods accordingly. The main objective of this paper is to highlight a few of these challenges and discuss the research opportunities in this direction.

\begin{figure}[!t]
\begin{tabular}{@{\hspace{0em}}l@{\hspace{0em}}l}
   \raisebox{3em}{
   \begin{turn}{90}
   UADFV
   \end{turn}
   }
   &\includegraphics[width=.45\textwidth]{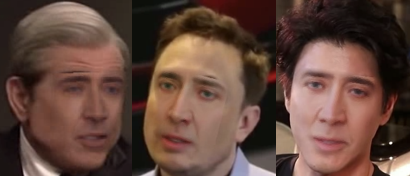} \\
   \raisebox{2em}{
   \begin{turn}{90}
   DF-TIMIT-HQ
   \end{turn}
   }
    &\includegraphics[width=.45\textwidth]{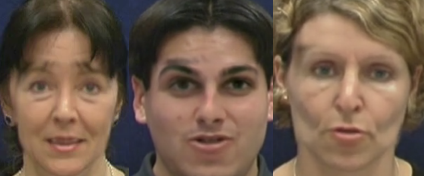} \\
    \raisebox{3em}{
   \begin{turn}{90}
   FF-DF
   \end{turn}
   }
    &\includegraphics[width=.45\textwidth]{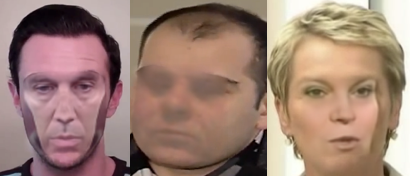} \\
    \raisebox{4em}{
   \begin{turn}{90}
   DFD
   \end{turn}
   }&\includegraphics[width=.45\textwidth]{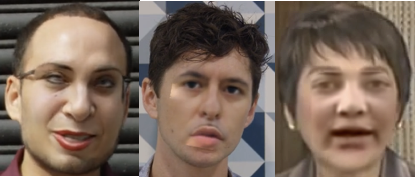} \\
    \raisebox{3em}{
   \begin{turn}{90}
   DFDC
   \end{turn}
   }&\includegraphics[width=.45\textwidth]{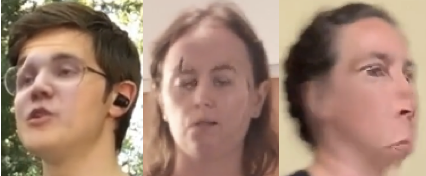}
\end{tabular}
~\vspace{-1em}
\caption{\em \small Visual artifacts of DeepFake videos in existing datasets, including low-quality, visible splicing boundaries, color mismatch, visible parts of the original face, and inconsistent face orientations. }
    \label{fig:overview}
    \vspace{-2em}
\end{figure}

\section{Current DeepFake Detection Methods} 

Current DeepFake detection methods mostly target face-swapping videos, which account for the majorities of DeepFake videos circulated online. Many of the existing methods are formulated as frame-level binary classification problems. Based on the features that are used, these methods fall into three major categories. Methods in the first category are based on inconsistencies exhibited in the {\bf physical/physiological} aspects in the DeepFake videos. The method in work of \cite{li2018ictu} exploits the observation that many DeepFake videos lack reasonable eye blinking due to the use of online portraits as training data, which usually do not have closed eyes for aesthetic reasons. Incoherent head poses in DeepFake videos are utilized in \cite{yang2018exposing} to expose DeepFake videos.  In~\cite{agarwal2019protecting}, the idiosyncratic behavioral patterns of a particular individual are captured by the time series of facial landmarks extracted from real videos are used to spot DeepFake videos. The second category of DeepFake detection algorithms (\eg, \cite{matern2019exploiting,li2019exposing}) use {\bf signal-level} artifacts introduced during the synthesis process. Also, as synthesized faces are spliced into the original video frames, state-of-the-art DNN splicing detection methods, \eg,~\cite{zhou2017two,zhou2018learning,liu2018image,bappy2019hybrid}, can be applied. The third category of DeepFake detection methods (\eg, \cite{afchar2018mesonet,guera2018deepfake,nguyen2019capsule,nguyen2019capsulev2}) are {\bf data-driven}, which directly employ various types of DNNs trained on real and DeepFake videos but capturing specific artifact. 

\subsection{Limitations}

Albeit impressive progress has been made in the performance of detection of DeepFake videos, there are several concerns over the current detection methods that suggest caution. 

\noindent{\bf Quality of DeepFake Datasets}. The availability of large-scale datasets of DeepFake videos is an enabling factor in the development of DeepFake detection method. However, a closer look at the DeepFake videos in existing datasets reveals some stark contrasts in visual quality to the actual DeepFake videos circulated on the Internet. Several common visual artifacts that can be found in these datasets are highlighted in Fig.\ref{fig:overview}, including low-quality synthesized faces, visible splicing boundaries, color mismatch, visible parts of the original face, and inconsistent synthesized face orientations. These artifacts are likely the result of imperfect steps of the synthesis method and the lack of curating of the synthesized videos before included in the datasets. Moreover, DeepFake videos with such low visual qualities can hardly be convincing, and are unlikely to have real impact. Correspondingly, high detection performance on these dataset may not bear strong relevance when the detection methods are deployed {\em in the wild}. A related issue is that DeepFake detection methods trained using different DF datasets have trouble extending the performance to different datasets \cite{li_etal_cvpr20}. 

In a recent work \cite{li_etal_cvpr20}, we present a new large-scale challenging DeepFake video dataset, {\em Celeb-DF}, which contains {$5,639$} high-quality DeepFake videos of celebrities generated using improved synthesis process.  We conduct a comprehensive evaluation of DeepFake detection methods and datasets to demonstrate the escalated level of challenges.
\begin{figure*}[t]
    \centering
    \includegraphics[width=\linewidth]{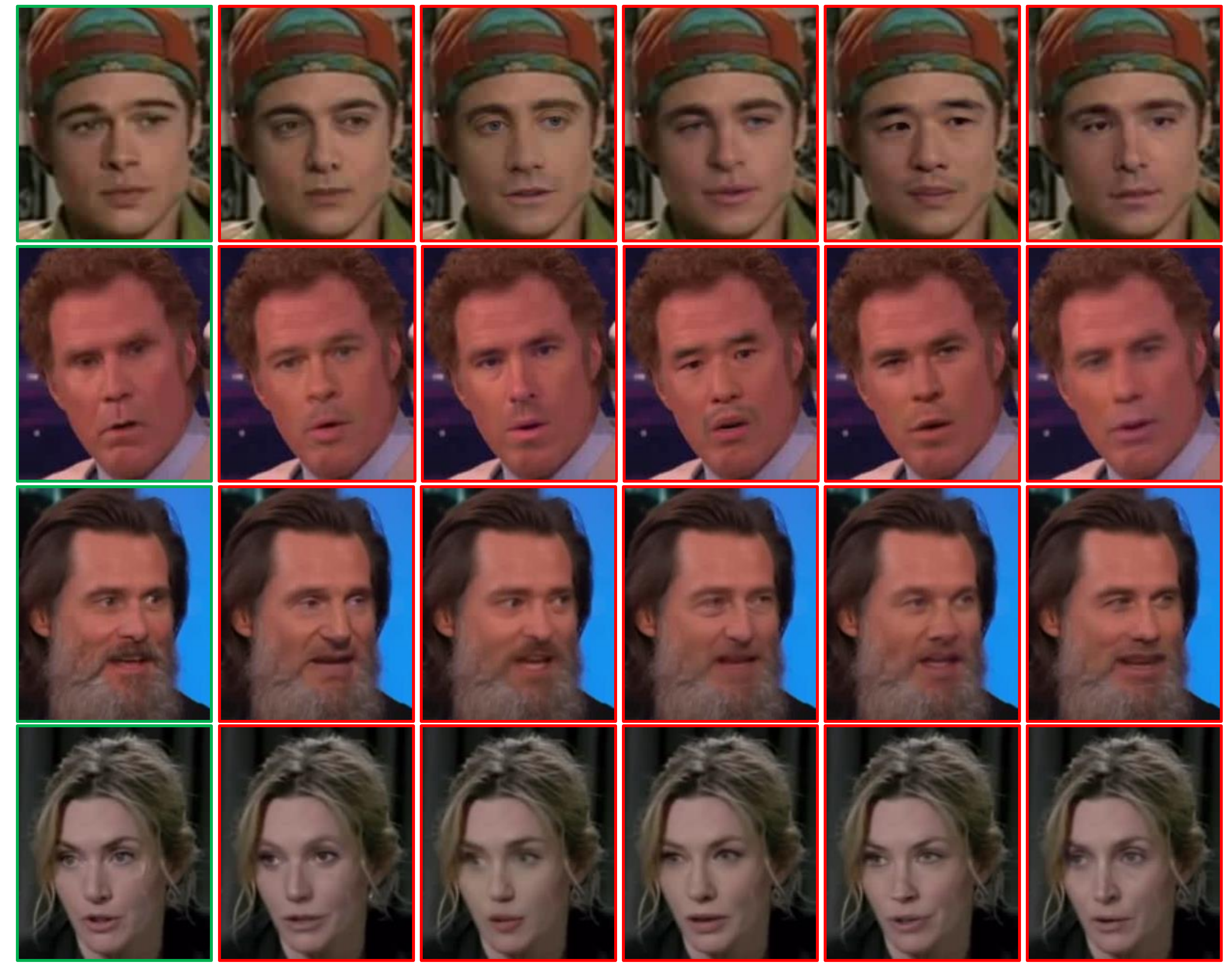}
        \vspace{-2em}
    \caption{\em \small Example frames from the Celeb-DF dataset. Left column is the frame of real videos and right five columns are corresponding DeepFake frames generated using different donor subject.}
    \label{fig:demo}
    \vspace{-1em}
\end{figure*}

\noindent{\bf Performance Evaluation}. Currently, the problem of detecting DeepFake videos is commonly formulated, solved, and evaluated as a binary classification problem, where each video is categorized as real or a DeepFake. Such dichotomy is easy to set up in controlled experiments, where we develop and test DeepFake detection algorithms using videos that are either pristine or made with DeepFake generation algorithms. However, the picture is murkier when the detection method is deployed in real world.  For instance, videos can be fabricated or manipulated in ways other than DeepFakes, so not being detected as a DeepFake video does not  necessarily suggest the video is a real one. Also, a DeepFake video may be subject to other types of manipulations and a single label may not comprehensively reflect such. Furthermore, in a video with multiple subjects' faces only one or a few are generated with DeepFake for a fraction of the frames. So the binary classification scheme needs to be extended to multi-class, multi-label, and local classification/detection to fully handle the complexities of real world media forgeries. 

\noindent{\bf Explainability of Detection Results}. Current DeepFake detection methods are usually designed to perform batch analysis over a large collection videos. However, when the detection methods are used in the field by journalists or law enforcement, we usually need only to analyze a small number of videos. Numerical score corresponding to the likelihood of a video being generated using a synthesis algorithm is not as useful to the practitioners if it is not corroborated with proper reasoning of the score. In such scenarios, it is very typical to request a justification for the numerical score for the analysis to be acceptable for publishing or used in court. However, many data-driven DF detection methods, especially those based on the use of deep neural networks, usually lack explainability due to the black box nature of the DNN models. 
%Lack of relevant and proper evaluation protocols and metrics: Performance metrics such as AUC, mAP, or EER are borrowed from problems in Computer vision, such as image classification or object detection. analysis put more emphasis on the explainability of the result while batch analysis concerns more on the scalability of the detection method. 

\noindent{\bf Temporal Aggregation}. Most existing DeepFake detection methods are based on binary classification at the frame level, \ie, determining the likelihood of an individual frame as real or of DeepFake. Although simple and straightforward, there are two issues of this methodology. First, the temporal consistency among frames are not explicitly considered, as (i) many DeepFake videos exhibit temporal artifacts and (ii) real or DeepFake frames tend to appear in continuous intervals. Second, it necessitates an extra step when video-level integrity score is needed: we have to aggregate the scores over individual frames to compute such a score. 

\noindent{\bf Social Media Laundering}. A large fraction of online videos are now spread through social networks, \eg, FaceBook, Instagram, and Twitter. To save network bandwidth and also to protect the users' privacy, these videos are usually striped off meta-data, down-sized, and then heavy compressed before they are uploaded to the social platforms. These operations, commonly known as {\em social media laundering}, are detrimental to recover traces of underlying manipulation, and at the same time increase the false positive detections, \ie, classifying a real video as a DeepFake. So far, most data-driven DeepFake detection methods that use signal level features are much affected by social media laundering. A practical measure to improve the robustness of DeepFake detection methods to social media laundering is to actively incorporate simulations of such effects in training data, and also enhance evaluation datasets to include performance on social media laundered videos, both real and synthesized.

\section{Future Directions}

Besides continuing improving to solve the aforementioned limitations, we also envision a few important directions of DeepFake detection methods that will receive more attention in the coming years. 

\noindent{\bf Other Forms of DeepFakes}. Although face swapping is currently the most widely known form of DeepFake videos, it is by no means the most effective. In particular, for the purpose of impersonating someone, face swapping DeepFake videos have several limitations. Psychological studies [citation] show that human face recognition largely relied on information gleaned from face shape and hairstyle. As such, to create convincing impersonating effect, the person whose face is to be replaced (the target) has to have similar face shape and hairstyle to the person whose face is used for swapping (the donor). Second, as the synthesized faces need to be spliced into the original video frame, the inconsistencies between the synthesized region and the rest of the original frame can be severe and difficult to conceal. 

In these respects, the other two forms of DeepFake videos, namely, head puppetry and lip-syncing, are more effective and thus should become the focus of subsequent research in DeepFake detection. Methods studying whole face synthesis or reenactment have experienced fast development in recent years. Although there have not been as many easy-to-use and free open-source software tools generating these types of DeepFake videos as for the face-swapping videos, the continuing sophistication of the generation algorithms will change the situation in the near future. Because the synthesized region is different from face swapping DeepFake videos (the whole face in the former and lip area in the latter), detection methods designed based on artifacts specific to face swapping are unlikely to be effective for these videos. Correspondingly, we should develop detection methods that are effective to these types of DeepFake videos. 

\noindent{\bf Audio DeepFakes}. AI-based impersonation are not limited to imagery, recent AI-synthesized content-generation are leading to the creation of highly realistic audios~\cite{ping2017deep,yu2018interspeech}. Using synthesized audios of the impersonating target can significantly make the DeepFake videos more convincing and compounds its negative impact. As audio signals are 1D signals and have very different nature from images and videos, different methods need to be developed to specifically targeting such forgeries. This problem has drawn attention in the speech processing community recently with part of the most recent Global ASVspoofing Challenge\footnote{\url{https://www.asvspoof.org/}.} dedicated to AI-driven voice conversion detection, and a few dedicated  methods for audio DeepFake detection, \eg, \cite{albadway_cvprw19a}, have also shown up recently. In the coming years, we expect more developments in these areas, in particular, those can leverage features in both visual and audio features of the fake videos. 

\noindent{\bf Intent Inference}. Even though the potential negative impacts of DeepFake videos are tremendous, in reality, the majority of DeepFake videos are not created not with a malicious intent. Many DeepFake videos currently circulated online are of a pranksome, humorous, or satirical nature. As such, it is important to expose the underlying intent of a DeepFake in the context of legal or journalistic investigation. Inferring intention may require more semantic and contextual understanding of the content, few forensic methods are designed to answer this question, but this is certainly a direction that future forensic methods will focus on.

\noindent{\bf Anti-forensics}. With the increasing effectiveness of DeepFake detection methods, we also anticipate developments of corresponding anti-forensic measures, which take advantage of the vulnerabilities of current DeepFake detection methods to conceal revealing traces of DeepFake videos. The data-driven deep neural network based DeepFake detection methods are particularly susceptible to anti-forensic attacks due to the known vulnerability of general deep neural network classification models. Anti-forensic measures can also be developed in the other aspect, to disguise a real video as a DeepFake video by adding simulated signal level features used by current detection algorithms, a situation we term as {\em fake DeepFake}. Further DeepFake detection methods must improve to handle such intentional and adversarial attacks.
\begin{figure}[t]
	\centering
	\includegraphics[width=0.9\linewidth]{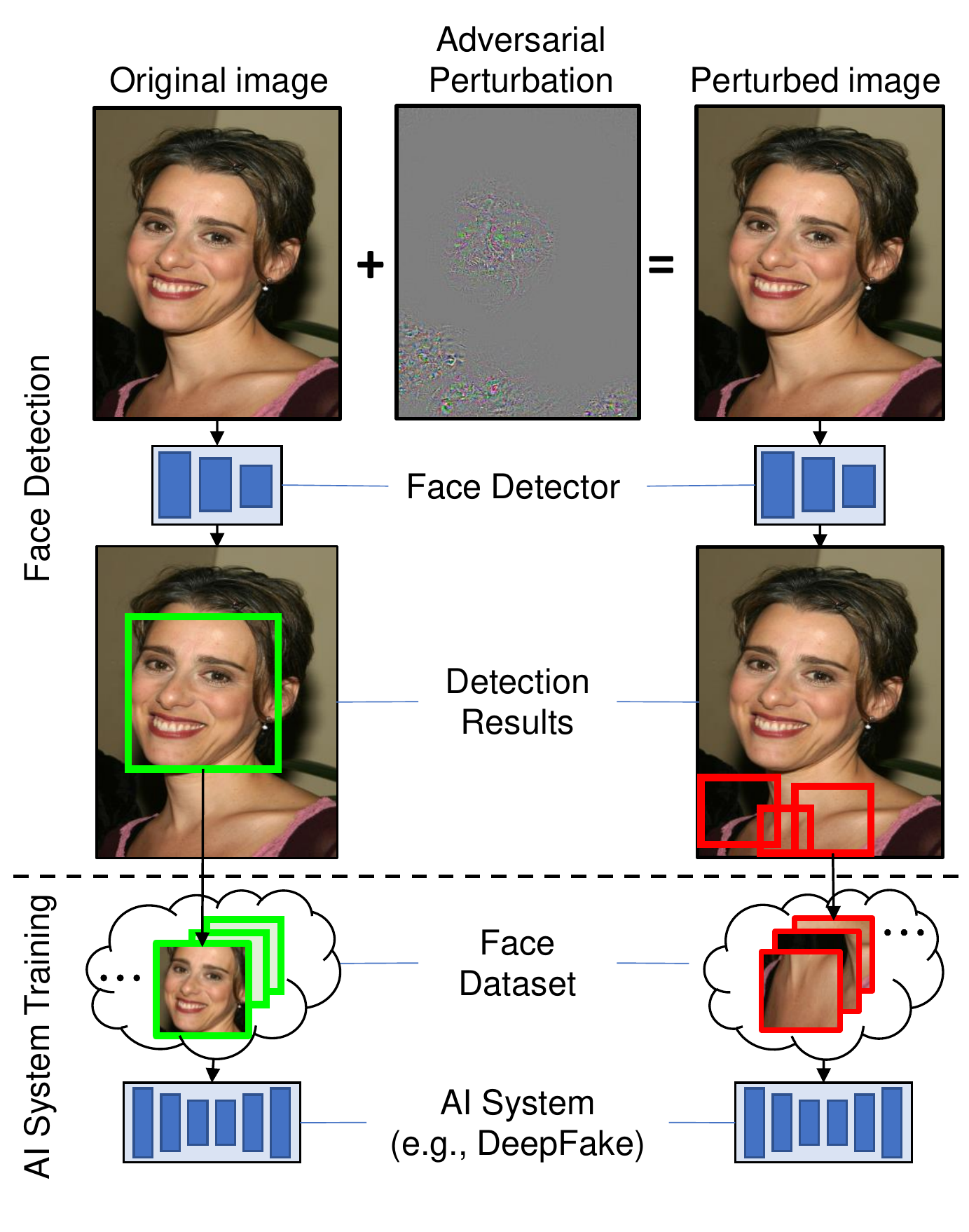}
	\vspace{-0.5cm}
	\caption{\em \small Overview of the proposed method of disrupting AI face synthesis. Our aim is to use the adversarial perturbations (amplified by $30$ for better visualization) to distract DNN-based face detectors, such that the quality of the obtained face set as training data to the AI face synthesis is reduced.}
	\label{fig:overview}
	~\vspace{-2.5em}
\end{figure}

\noindent{\bf Human Performance}. Although the potential negative impacts of online DeepFake videos are widely recognized, currently there is a lack of formal and quantitative study of the perceptual and psychological factors underlying their deceptiveness. Interesting questions such as if there exist an {\em uncanny valley}\footnote{The uncanny valley in ths context refers to the phenomenon whereby a DeepFake generated face bearing a near-identical resemblance to a human being arouses a sense of unease or revulsion in the viewers.} for DF videos, what is the {\em just noticeable difference} between high-quality DeepFake videos and real videos to human eyes, or what type/aspects of DeepFake videos are more effective in deceiving the viewers, have yet to be answered. To pursue these questions, it calls for close collaboration among  researchers in digital media forensics and in perceptual and social psychology. There is no doubt that such studies are invaluable to research in detection techniques as well as a better understanding of the social impact that DeepFakes can cause.

%%%%%%%%%%%%%%%%%%%

%signal based DF detection can be unreliable. The majority of DF detection algorithms actually do not answer the question ``if this video is DF’’ directly, but rather an indirect question if the signal, physical, or physiological level cues that have been observed in training DF videos are present in this video’’. This decision gap can spawn some confusions, and will also make the DF detection algorithms vulnerable to attacks. 

\noindent{\bf Protection measures}. However, given the speed and reach of the propagation of online media, even the currently best forensic techniques will largely operate in a postmortem fashion, applicable only after AI synthesized fake face images or videos emerge. We aim to develop {\em proactive} approaches to protect individuals from becoming the victims of such attacks, which complement to the forensic tools. 

One such method we have recently studied \cite{li2019hiding} is to add specially designed patterns known as the {\em adversarial perturbations} that are imperceptible to human eyes but can result in detection failures. The rationale is as follows. High-quality AI face synthesis models need large number of, typically in the range of thousands, sometimes even millions, training face images collected using automatic face detection methods, \ie, the {\em face sets}. Adversarial perturbations ``pollute'' a face set to have few actual faces and many non-faces with low or no utility as training data for AI face synthesis models, Fig. \ref{fig:overview}. The proposed adversarial perturbation generation method can be implemented as a service of photo/video sharing platforms before a user's personal images/videos are uploaded or as a standalone tool that the user can use, to process the images and videos before they are uploaded online.

%While in real world, we know that the fake images or videos may be the result of multiple operations applied sequentially, and some may be unknown to us in training. How to make the forensic detection methods flexible in such complex scenarios will be an important topic for future digital media forensics. 

\section{Conclusion}

We predict that several future technological developments will further improve the visual quality and generation efficiency of the fake videos. Firstly, one critical disadvantage of the current DeepFake generation methods are that they cannot produce good details such as skin and facial hairs. This is due to the loss of information in the encoding step of generation. However, this can be improved by incorporating GAN models\cite{gan_goodfellow_nips_2014} which have demonstrated performance in recovering facial details in recent works \cite{gan_karras_iclr_2017,gan_karras_cvpr_2019}. Secondly, the synthesized videos can be more realistic if they are accompanied with realistic voices, which combines video and audio synthesis together in one tool. 

In the face of this, the overall running efficiency, detection accuracy, and more importantly, false positive rate, have to be improved for wide practical adoption.  The detection methods also need to be more robust to real-life post-processing steps, social media laundering, and counter-forensic technologies.  There is a perpetual competition of technology, know-hows, and skills between the forgery makers and digital media forensic researchers. The future will reckon the predictions we make in this work.

%Last but not least, not merely a technical problem. We need a more close collaboration from the government agencies, platform companies, media outlets, technical and research community, as well as the ordinary online users. 

{\small
\bibliographystyle{IEEEbib}
\bibliography{refs,egbib,others,lyu}

\begin{thebibliography}{1}

\bibitem{Authors12}
Authors,
\newblock ``The frobnicatable foo filter,'' ACM MM 2013 submission ID 324.
  Supplied as additional material {\tt acmmm13.pdf}.

\bibitem{Authors12b}
Authors,
\newblock ``Frobnication tutorial,'' 2012,
\newblock Supplied as additional material {\tt tr.pdf}.

\bibitem{Morgan2005}
Dennis~R. Morgan,
\newblock ``Dos and don'ts of technical writing,''
\newblock {\em IEEE Potentials}, vol. 24, no. 3, pp. 22--25, Aug. 2005.

\bibitem{cooley65}
J.~W. Cooley and J.~W. Tukey,
\newblock ``An algorithm for the machine computation of complex {F}ourier
  series,''
\newblock {\em Math. Comp.}, vol. 19, pp. 297--301, Apr. 1965.

\bibitem{haykin02}
S.~Haykin,
\newblock ``Adaptive filter theory,''
\newblock Information and System Science. Prentice Hall, 4th edition, 2002.

\end{thebibliography}


\begin{thebibliography}{10}

\bibitem{farid08}
Hany Farid,
\newblock {\em Digital Image Forensics},
\newblock MIT Press, 2012.

\bibitem{fakeapp}
``{F}ake{A}pp,'' \url{https://www.malavida.com/en/soft/fakeapp/}, Acessed Nov
  4, 2019.

\bibitem{DFaker}
``{DF}aker github,'' \url{https://github.com/dfaker/df}, Accessed Nov 4, 2019.

\bibitem{faceswap-gan}
``faceswap-{GAN} github,'' \url{https://github.com/shaoanlu/faceswap-GAN},
  Accessed Nov 4, 2019.

\bibitem{faceswap}
``faceswap github,'' \url{https://github.com/deepfakes/faceswap}, Accessed Nov
  4, 2019.

\bibitem{DeepFaceLab}
``{D}eep{F}ace{L}ab github,'' \url{https://github.com/iperov/DeepFaceLab},
  Accessed Nov 4, 2019.

\bibitem{survey_chesney_citron_2018}
Robert Chesney and Danielle~Keats Citron,
\newblock ``{Deep Fakes: A Looming Challenge for Privacy, Democracy, and
  National Security},''
\newblock {\em 107 California Law Review (2019, Forthcoming); U of Texas Law,
  Public Law Research Paper No. 692; U of Maryland Legal Studies Research Paper
  No. 2018-21}.

\bibitem{afchar2018mesonet}
Darius Afchar, Vincent Nozick, Junichi Yamagishi, and Isao Echizen,
\newblock ``Mesonet: a compact facial video forgery detection network,''
\newblock in {\em IEEE International Workshop on Information Forensics and
  Security (WIFS)}, 2018.

\bibitem{guera2018deepfake}
David G{\"u}era and Edward~J Delp,
\newblock ``Deepfake video detection using recurrent neural networks,''
\newblock in {\em AVSS}, 2018.

\bibitem{li2018ictu}
Yuezun Li, Ming-Ching Chang, and Siwei Lyu,
\newblock ``In ictu oculi: Exposing {AI} generated fake face videos by
  detecting eye blinking,''
\newblock in {\em IEEE International Workshop on Information Forensics and
  Security (WIFS)}, 2018.

\bibitem{yang2018exposing}
Xin Yang, Yuezun Li, and Siwei Lyu,
\newblock ``Exposing deep fakes using inconsistent head poses,''
\newblock in {\em IEEE International Conference on Acoustics, Speech and Signal
  Processing (ICASSP)}, 2019.

\bibitem{matern2019exploiting}
Falko Matern, Christian Riess, and Marc Stamminger,
\newblock ``Exploiting visual artifacts to expose deepfakes and face
  manipulations,''
\newblock in {\em IEEE Winter Applications of Computer Vision Workshops
  (WACVW)}, 2019.

\bibitem{li2019exposing}
Yuezun Li and Siwei Lyu,
\newblock ``Exposing deepfake videos by detecting face warping artifacts,''
\newblock in {\em IEEE Conference on Computer Vision and Pattern Recognition
  Workshops (CVPRW)}, 2019.

\bibitem{sabir2019recurrent}
Ekraam Sabir, Jiaxin Cheng, Ayush Jaiswal, Wael AbdAlmageed, Iacopo Masi, and
  Prem Natarajan,
\newblock ``Recurrent-convolution approach to deepfake detection-state-of-art
  results on faceforensics++,''
\newblock {\em arXiv preprint arXiv:1905.00582}, 2019.

\bibitem{roessler2019faceforensics++}
Andreas R\"ossler, Davide Cozzolino, Luisa Verdoliva, Christian Riess, Justus
  Thies, and Matthias Nie{\ss}ner,
\newblock ``Face{F}orensics++: Learning to detect manipulated facial images,''
\newblock in {\em ICCV}, 2019.

\bibitem{nguyen2019capsule}
Huy~H Nguyen, Junichi Yamagishi, and Isao Echizen,
\newblock ``Capsule-forensics: Using capsule networks to detect forged images
  and videos,''
\newblock in {\em IEEE International Conference on Acoustics, Speech and Signal
  Processing (ICASSP)}, 2019.

\bibitem{nguyen2019multi}
Huy~H Nguyen, Fuming Fang, Junichi Yamagishi, and Isao Echizen,
\newblock ``Multi-task learning for detecting and segmenting manipulated facial
  images and videos,''
\newblock in {\em IEEE International Conference on Biometrics: Theory,
  Applications and Systems (BTAS)}, 2019.

\bibitem{nguyen2019capsulev2}
Huy~H Nguyen, Junichi Yamagishi, and Isao Echizen,
\newblock ``Use of a capsule network to detect fake images and videos,''
\newblock {\em arXiv preprint arXiv:1910.12467}, 2019.

\bibitem{korshunov2018deepfakes}
Pavel Korshunov and S{\'e}bastien Marcel,
\newblock ``Deepfakes: a new threat to face recognition? assessment and
  detection,''
\newblock {\em arXiv preprint arXiv:1812.08685}, 2018.

\bibitem{DDD_GoogleJigSaw2019}
Nicholas Dufour, Andrew Gully, Per Karlsson, Alexey~Victor Vorbyov, Thomas
  Leung, Jeremiah Childs, and Christoph Bregler,
\newblock ``Deepfakes detection dataset by google \& jigsaw,'' .

\bibitem{dolhansky2019deepfake}
Brian Dolhansky, Russ Howes, Ben Pflaum, Nicole Baram, and Cristian~Canton
  Ferrer,
\newblock ``The deepfake detection challenge ({DFDC}) preview dataset,''
\newblock {\em arXiv preprint arXiv:1910.08854}, 2019.

\bibitem{agarwal2019protecting}
Shruti Agarwal, Hany Farid, Yuming Gu, Mingming He, Koki Nagano, and Hao Li,
\newblock ``Protecting world leaders against deep fakes,''
\newblock in {\em IEEE Conference on Computer Vision and Pattern Recognition
  Workshops (CVPRW)}, 2019.

\bibitem{zhou2017two}
Peng Zhou, Xintong Han, Vlad~I Morariu, and Larry~S Davis,
\newblock ``Two-stream neural networks for tampered face detection,''
\newblock in {\em IEEE Conference on Computer Vision and Pattern Recognition
  Workshops (CVPRW)}, 2017.

\bibitem{zhou2018learning}
Peng Zhou, Xintong Han, Vlad~I Morariu, and Larry~S Davis,
\newblock ``Learning rich features for image manipulation detection,''
\newblock in {\em CVPR}, 2018.

\bibitem{liu2018image}
Yaqi Liu, Qingxiao Guan, Xianfeng Zhao, and Yun Cao,
\newblock ``Image forgery localization based on multi-scale convolutional
  neural networks,''
\newblock in {\em ACM Workshop on Information Hiding and Multimedia Security
  (IHMMSec)}, 2018.

\bibitem{bappy2019hybrid}
Jawadul~H Bappy, Cody Simons, Lakshmanan Nataraj, BS~Manjunath, and Amit~K
  Roy-Chowdhury,
\newblock ``Hybrid lstm and encoder-decoder architecture for detection of image
  forgeries,''
\newblock {\em IEEE Transactions on Image Processing (TIP)}, 2019.

\bibitem{li_etal_cvpr20}
Yuezun Li, Pu~Sun, Honggang Qi, and Siwei Lyu,
\newblock ``{Celeb-DF: A Large-scale Challenging Dataset for DeepFake
  Forensics},''
\newblock in {\em IEEE Conference on Computer Vision and Patten Recognition
  (CVPR)}, Seattle, WA, United States, 2020.

\bibitem{ping2017deep}
Wei Ping, Kainan Peng, Andrew Gibiansky, Sercan~O Arik, Ajay Kannan, Sharan
  Narang, Jonathan Raiman, and John Miller,
\newblock ``Deep voice 3: 2000-speaker neural text-to-speech,''
\newblock {\em arXiv preprint arXiv:1710.07654}, 2017.

\bibitem{yu2018interspeech}
Yu~Gu and Yongguo Kang,
\newblock ``Multi-task {WaveNet}: A multi-task generative model for statistical
  parametric speech synthesis without fundamental frequency conditions,''
\newblock in {\em Interspeech}, Hyderabad, India, 2018.

\bibitem{albadway_cvprw19a}
Ehab AlBadawy, Siwei Lyu, and Hany Farid,
\newblock ``Detecting ai-synthesized speech using bispectral analysis,''
\newblock in {\em Workshop on Media Forensics (in conjunction with CVPR)}, Long
  Beach, CA, United States, 2019.

\bibitem{li2019hiding}
Yuezun Li, Xin Yang, Baoyuan Wu, and Siwei Lyu,
\newblock ``Hiding faces in plain sight: Disrupting ai face synthesis with
  adversarial perturbations,'' 2019.

\bibitem{gan_goodfellow_nips_2014}
Ian Goodfellow, Jean Pouget-Abadie, Mehdi Mirza, Bing Xu, David Warde-Farley,
  Sherjil Ozair, Aaron Courville, and Yoshua Bengio,
\newblock ``Generative adversarial nets,''
\newblock in {\em Advances in Neural Information Processing Systems (NIPS)},
  2014, pp. 2672--2680.

\bibitem{gan_karras_iclr_2017}
Tero Karras, Timo Aila, Samuli Laine, and Jaakko Lehtinen,
\newblock ``Progressive growing of gans for improved quality, stability, and
  variation,''
\newblock {\em The International Conference on Learning Representations
  (ICLR)}, 2017.

\bibitem{gan_karras_cvpr_2019}
Tero Karras, Samuli Laine, and Timo Aila,
\newblock ``A style-based generator architecture for generative adversarial
  networks,''
\newblock in {\em Proceedings of the IEEE Conference on Computer Vision and
  Pattern Recognition}, 2019, pp. 4401--4410.

\end{thebibliography}
}

\end{document}